\documentclass[10pt,twocolumn,letterpaper]{article}

\usepackage{cvpr}
\usepackage{times}
\usepackage{epsfig}
\usepackage{graphicx}
\usepackage{subfigure}
\usepackage{amsmath}
\usepackage{amssymb}
\usepackage{mathtools}
\usepackage{color}
\usepackage[dvipsnames]{xcolor}

\usepackage[pagebackref=true,breaklinks=true,letterpaper=true,colorlinks,bookmarks=false]{hyperref}
\hypersetup{linkcolor=[rgb]{0.8551,0.2333,0.2333}}
\hypersetup{citecolor=[rgb]{0.3333,0.2333,0.7551}}

\cvprfinalcopy 

\def\cvprPaperID{2742} 

\ifcvprfinal\fi
\begin{document}

\title{Learning Strict Identity Mappings in Deep Residual Networks}

\author{
Xin Yu$^{1}$ \quad Zhiding Yu$^{2}$ \quad Srikumar Ramalingam$^{1}$
\\$^1$ University of Utah \quad $^2$ NVIDIA
\\{\tt\small \{xiny,srikumar\}@cs.utah.com, zhidingy@nvidia.com
}
}

\maketitle

\begin{abstract} A family of super deep networks, referred to as residual networks or ResNet~\cite{he2016deep}, achieved record-beating performance in various visual tasks such as image recognition, object detection, and semantic segmentation. The ability to train very deep networks naturally pushed the researchers to use enormous resources to achieve the best performance. Consequently, in many applications super deep residual networks were employed for just a marginal improvement in performance. In this paper, we propose $\epsilon$-ResNet that allows us to automatically discard redundant layers, which produces responses that are smaller than a threshold $\epsilon$, with a marginal or no loss in performance. The $\epsilon$-ResNet architecture can be achieved using a few additional rectified linear units in the original ResNet. Our method does not use any additional variables nor numerous trials like other hyper-parameter optimization techniques. The layer selection is achieved using a single training process and the evaluation is performed on CIFAR-10, CIFAR-100, SVHN, and ImageNet datasets. In some instances, we achieve about 80\% reduction in the number of parameters.
\end{abstract}

\section{Introduction}

\begin{figure}[t]
\begin{center}
\includegraphics[width=0.47\textwidth]{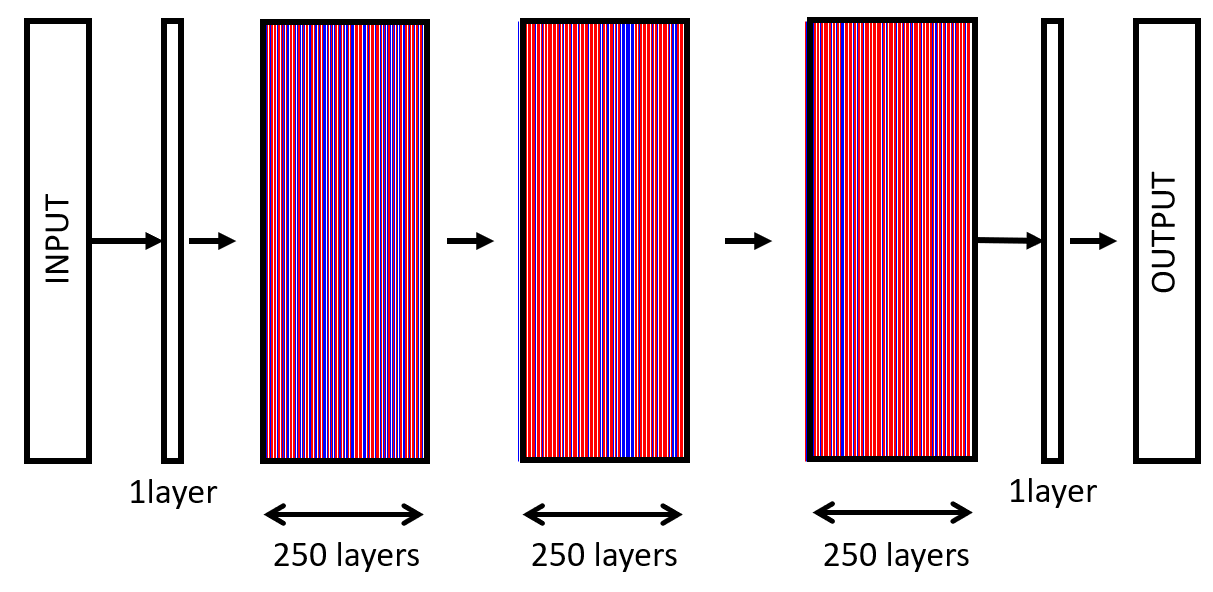}
\end{center}
\caption{We show a very deep $\epsilon$-residual network with 752 layers for training CIFAR-100~\cite{krizhevsky2009learning}. During training, $\epsilon$-ResNet identifies the layers that can be discarded with marginal or no loss in performance. The red lines indicate the layers that can be pruned, and the blue lines show the layers that need to be used. In this particular instance, we achieve a compression ratio of 3.2 (original number of layers / reduced number of layers). The validation errors of the original and the reduced networks are given by 24.8\% and 23.8\%, respectively.}
\label{intro_figure}
\end{figure}

The basic idea behind $\epsilon$-ResNet is shown in Fig.~\ref{intro_figure} where $\epsilon$-ResNet is trained on the CIFAR-100 dataset~\cite{krizhevsky2009learning}. In particular, we show a 752-layer network with each residual block having 2 convolution layers and the ``pre-activation'' setting following~\cite{he2016identity}. During the training, we automatically identify the layers that can be pruned or discarded without any loss (or with marginal loss) in the performance. We achieve this by modifying the standard residual network with a few additional rectified linear units that automatically discards residual blocks whose responses are below a threshold. In this particular instance shown, we achieve a compression ratio of around $3.2$ (original number of layers / reduced number of layers).

Recent advances in representation learning have demonstrated the powerful role played by deep residual learning~\cite{he2016deep}. As a result, ResNet has pushed the boundaries of a wide variety of vision tasks significantly, including but are not limited to general object recognition~\cite{he2016deep,xie2017aggregated}, object detection~\cite{ren2017faster, dai2016r, ren2017object}, face recognition~\cite{liu2017sphereface}, segmentation~\cite{chen2017deeplab, wu2016wider, zhao2017pyramid} and semantic boundary detection~\cite{yu2017casenet}. More recently, He et al.~\cite{he2016identity} proposed an improved design of residual unit, where identity mappings are constructed by viewing the activation functions as ``pre-activation'' of the weight layers, in contrast to the conventional ``post-activation'' manner. This further led to considerably improved performance on very deep network architectures, such as a 1001-layer ResNet.

\begin{figure}[t]
\begin{center}
\mbox{
\subfigure[]{\psfig{figure=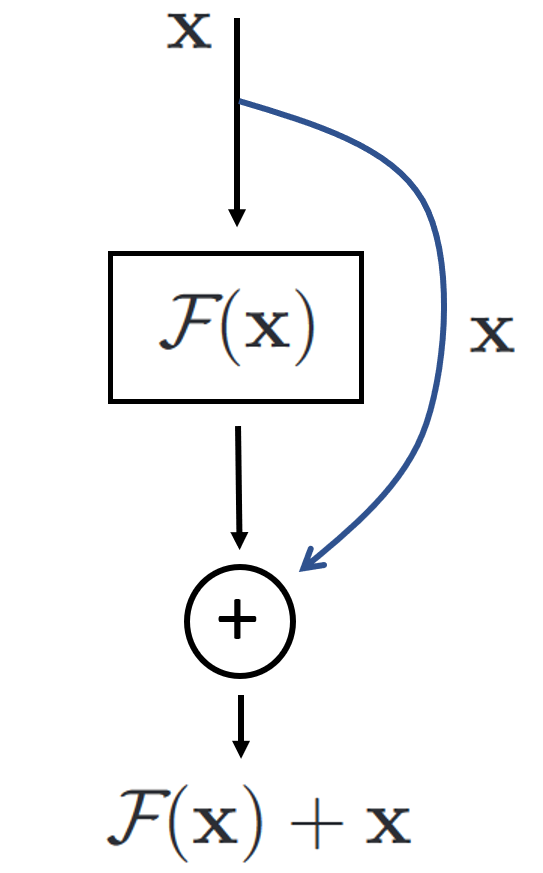,width=0.45\columnwidth}\label{fg.epsilon_resnet_idea.a}}
\subfigure[]{\psfig{figure=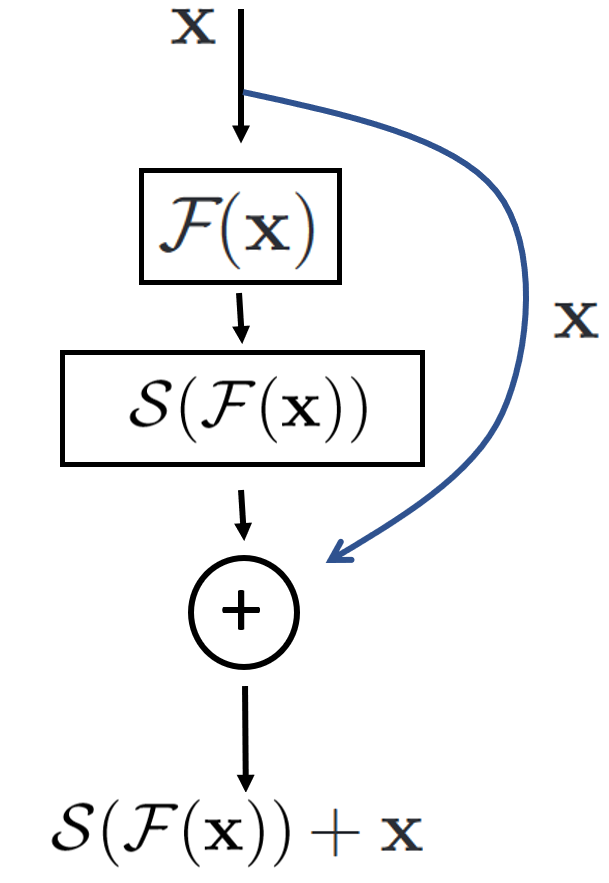,width=0.45\columnwidth}\label{fg.epsilon_resnet_idea.b}}
}
\end{center}
\caption{We show the basic transformation in standard ResNet and
    $\epsilon$-ResNet (a) The function mapping in standard ResNet is given by
        ${\cal H}({\bf x}) = {\cal F}(x) + x$. (b) The function mapping in
        $\epsilon$-ResNet is given by ${\cal H}({\bf x}) = {\cal S}({\cal F}(x))
        + x$. If all the residual responses in ${\cal F}(x)$ is less than a
        threshold $\epsilon$, then ${\cal S}({\cal F}(x)) = {\bf 0}$. If one of the responses are not small, then we do the same mapping ${\cal S}({\cal F}(x)) = {\cal F}(x)$ as the standard network.}
\label{fg.epsilon_resnet_idea}
\end{figure}

The remarkable success of ResNet leads to some obvious questions: What makes it work better than earlier architectures? One advantage with ResNet is its ability to handle vanishing/exploding gradients. However, the success could not be attributed only to this since many prior methods have already handled this with normalized initialization~\cite{Lecun1998,Glorot2010,He2015}. Another key contributing factor is the depth, which has been proven to be extremely beneficial in model expressiveness~\cite{Hastad1991,Montufar2014,Serra2017}. It was observed that training very deep neural networks is not a straightforward task as we encounter the ``under-fitting'' problem, where the training error keeps increasing with the depth of the network. This is in stark contrast of the natural expectation of ``over-fitting'' that typically happens when we use too many parameters. To illustrate this further, let us consider Fig.~\ref{fg.epsilon_resnet_idea.a}. Let ${\cal H}(x)$ be the desired underlying mapping, and we cast it as ${\cal F}(x) + x$. Here ${\cal F}(x) = {\cal H}(x) - x$ is the residual mapping. In \cite{he2016deep}, the following explanation is provided for the under-fitting problem:

\begin{quote}
\emph{``We hypothesize that it is easier to optimize the residual mapping than to optimize the original, unreferenced mapping.  To the extreme, if an identity mapping were optimal, it would be easier to push the residual to zero than to fit an identity mapping by a stack of nonlinear layers.''}
\end{quote}

While this makes sense, in reality one seldom observes the residuals going to perfect zeros or even negligible values in experiments with very deep networks. In this paper, we propose a technique that promotes zero residuals, which in other words achieves identity mapping in a strict sense.

\begin{figure}[t]
\begin{center}
\includegraphics[width=0.47\textwidth]{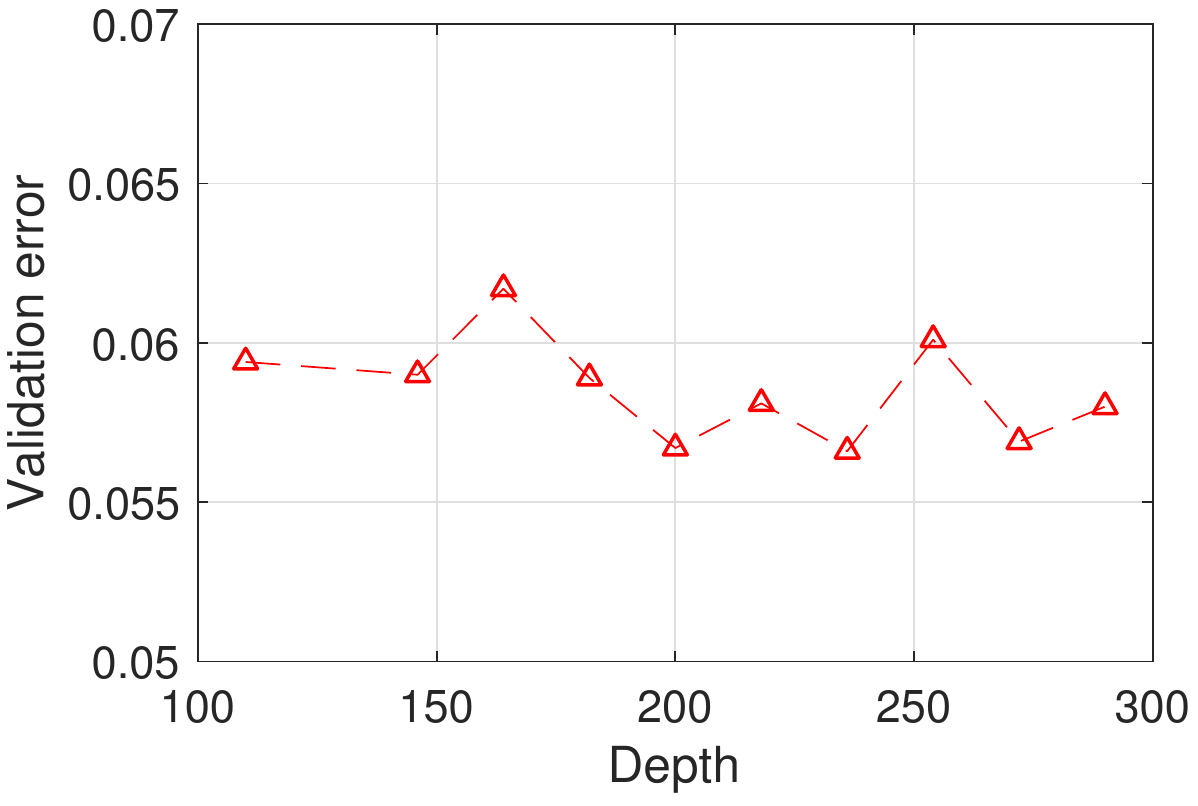}
\end{center}
\caption{validation errors of full pre-activation ResNet with different number of layers in CIFAR-10 dataset~\cite{krizhevsky2009learning}.}\label{fig1}
\end{figure}

A different interpretation for residual networks was given in~\cite{Veit2016}, where ResNet was treated as an ensemble of many shallow networks. In particular, they show several interesting experiments to demonstrate the role of depth and layers in VGGNet and ResNet. In particular, they show the deletion of one layer in VGGNet can lead to 80\% increase in error, while one barely notices the difference in ResNet. This shows that depth alone may not be the single key factor in the success of ResNet. Their hypothesis is that the multiple short and sometimes redundant paths play an important role in the performance. In order to study the role of depth in residual networks, we performed a simple experiment with CIFAR-10 dataset~\cite{krizhevsky2009learning}. We trained a family of deep residual networks with monotonically increasing number of layers from 100 to 300 as shown in Fig.~\ref{fig1}. As we observe, the error on the validation set is not monotonically decreasing. For example, the validation errors of 254-layer and 200-layer networks are given by 5.96\% and 5.66\%. Although there is a overall decrease in the error rate as we increase the number of layers, the behaviour is not strictly monotonic. Therefore, we would like to ask the following questions:
\begin{quote}
\emph{By training a residual network ${\cal N}$ with $n$ layers, can we find a reduced network ${\cal N}_R$ with $m \ll n$ layers without significant performance loss?}
\end{quote}

In this paper we propose $\epsilon$-ResNet, a variant of standard residual
networks, that promotes strict identity mapping. We illustrate the basic idea
behind $\epsilon$-ResNet in Fig.~\ref{fg.epsilon_resnet_idea.b}. We model the
desired unknown mapping as ${\cal H}({\bf x}) = {\cal S}({\cal F}({\bf x})) +
{\bf x}$. When all the responses in the original residual block ${\cal F}({\bf
        x})$ is below a threshold $\epsilon$, then we force ${\cal S}({\cal
            F}({\bf x}))={\bf 0}$. If any single response is not small, then we
        use the original mapping ${\cal S}({\cal F}({\bf x}))={\cal F}({\bf x})$
        as the standard residual network. In this paper, we will show that when a residual block produces zero responses using our proposed variant, the weights in the CNN filters of the corresponding residual block will be pushed to zeros by the training loss function that consists of cross-entropy term and $L2$ norm of the weight parameters with momentum optimization~\cite{sutskever2013importance}. Consequently, during the prediction or test time, we can safely remove these layers and build a reduced network. A direct benefit of such framework is model compression and faster inference when there is layer redundancy. In our experiments, $\epsilon$-ResNet produces remarkable reduction of the model size with marginal or no loss in performance. Our expectation is that $\epsilon$-ResNet should at least achieve good trade-off between performance and model size/complexity.

\section{Related Work}
\noindent
{\bf Residual networks and variants:} A significant direction of network design has been focusing on designing skip layer architectures in order to alleviate the vanishing/exploding gradient issues~\cite{srivastava2015highway}. More recent advances of this direction have led to the family of deep residual
networks~\cite{he2016deep, xie2017aggregated, szegedy2017inception}. He et al. further proposed an improved residual unit with full pre-activation~\cite{he2016identity}, where identity mappings are used as skip connections. Under this framework, it was mathematically shown that the feature of any deeper unit can be represented as the feature of any shallower unit plus the summation of the preceding residual responses. Such characteristic leads to significantly improved performance on ultra deep networks with more than 1000 layers, while previously proposed residual units suffer from over-fitting and degraded performance. However, saturation of performance gain still exists as the layer number increases, meaning that there is room for removing redundancy and achieve better trade-off between performance and model complexity.

\noindent
{\bf Network structure optimization:} The problem addressed in this paper can be seen as one of the subproblems of hyper-parameter optimization, where we identify a subnetwork by dropping unnecessary layers without any (or marginal) loss in the performance. We will briefly review the underlying ideas in hyper-parameter optimization. One of the distinctly unsolved problems in deep learning is the ability to automatically choose the right network architecture for solving a particular task. The popular ones (e.g., AlexNet, GoogLeNet, and Residual Networks) have shown record beating performance on various image recognition tasks. Nevertheless, it is a time-consuming process to decide on the hyper-parameters ({\sc HP}s) of a network such as the number of layers, type of activation functions, learning rate, and loss functions.

Non-practitioners of deep learning might ask: why is it difficult to optimize a few {\sc HP}s, while we already train millions of network weight parameters? One can view this as a global optimization of a black-box loss function $f$. The goal is to find the {\sc HP}s $\theta_{h}$ that minimizes $f(\theta_{h},\theta_{w},{\cal D}_{val})$, such that $\theta_{w} = \arg\min_{\theta_w}f(\theta_{h},\theta_{w},{\cal D}_{train})$.
Here $\theta_w$, ${\cal D}_{train}$, and ${\cal D}_{val}$ denote the weight parameters, training dataset, and validation dataset, respectively. Some of the {\sc HP}s such as the depth are discrete, and the topology of the network changes for various depth values. This makes it hard to treat both sets of parameters $\theta_{h}$ and $\theta_{w}$ in the same manner. Existing approaches evaluate the $f(\theta_{h},\theta_{w},{\cal D}_{val})$ for different values of $\theta_{h}$ and identify the optimal $\theta_{h}$.

Standard approaches include search strategies (manual, grid, and random) and Bayesian techniques. A grid search is a brute-force strategy that evaluates the function for all {\sc HP} values in a manually specified interval. In a random search, we evaluate on a random subset of parameter values to identify the optimum. Manual search denotes the process of greedily optimizing one parameter at a time, and then moving on to the next one -- we memorize and analyze our previous results, and this gives us an advantage over naive grid or random search. Practical considerations in the manual tuning of {\sc HP}s are provided for efficient training and debugging of large-scale networks~\cite{Bengio2012}.

Bayesian methods~\cite{Snoek2012,Bergstra2011} can automate the process of manual tuning by learning a statistical model of the function that maps the hyper-parameter values to the performance on the validation set. In other words, we can think of Bayesian methods as modeling the conditional probability $p(f|\theta_h)$ where $f$ is the performance on the validation set given {\sc HP}s $\theta_h$. By studying the performance of different search techniques on the 117 datasets from ~\cite{Feurer2015}, it was shown that many recent methods are only marginally better than random search~\cite{Recht2016}. It is highly recommended that all {\sc HP} optimization methods should be compared with random search baseline~\cite{Bergstra2012}. Reinforcement learning~\cite{Zoph2017} and evolutionary algorithms~\cite{Real2017} have also been used, but these methods use evaluations for a large number of parameter trials, and it is time-consuming for even small-scale problems. While many of these methods are driven towards finding the optimal network architecture that produces best performance, our work focuses on a subproblem where we identify a subnetwork that produces more-or-less the same results as the original one in a specific case of residual networks.

Many other researchers have looked at model compression using other techniques such as low-rank decomposition~\cite{denton2014exploiting, jaderberg2014speeding, zhang2015efficient}, quantization~\cite{rastegari2016xnor,courbariaux2016binarized,wu2016quantized}, architecture design~\cite{szegedy2015going, iandola2016squeezenet, howard2017mobilenets}, pruning~\cite{han2015learning,li2017pruning,molchanov2017pruning}, sparse~\cite{liu2015sparse,zhou2016less,alvarez2016learning,wen2016learning,adaptiveGraphs} learning, etc. Recently, sparse structure selection has been shown for residual networks using scaling factor variables~\cite{Huang2017}. These additional scaling variables are trained along with the standard weight parameters using $L1$ regularization terms using stochastic Accelerated Proximal Gradient (APG) method. While ~\cite{Huang2017} and $\epsilon$-ResNet share the same goal of discarding redundant layers, the techniques are entirely different. We do not add any additional variables to the standard residual networks. While \cite{Huang2017} uses $L1$ relaxation for solving $L0$ sparsity in the loss function, we promote layer sparsity by redesigning the network architecture that can achieve strict identity mapping.

\begin{figure*}[t]
\begin{center}
\mbox{
\subfigure[]{\psfig{figure=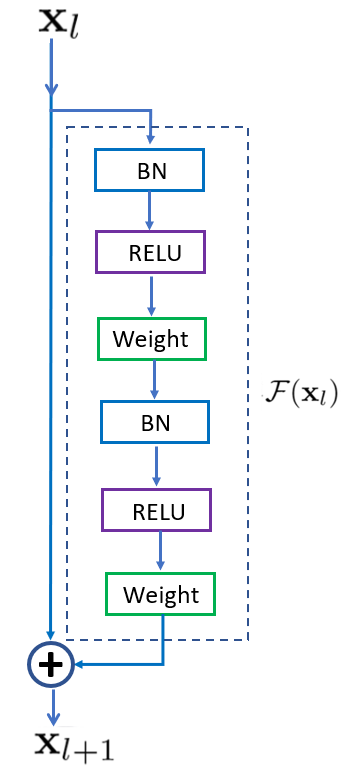,width=0.60\columnwidth}\label{fg.epsilon_resnet_design.a}}
\subfigure[]{\psfig{figure=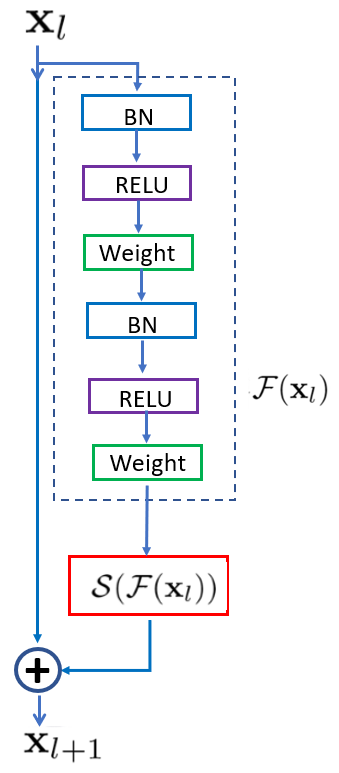,width=0.60\columnwidth}\label{fg.epsilon_resnet_design.b}}
\subfigure[]{\psfig{figure=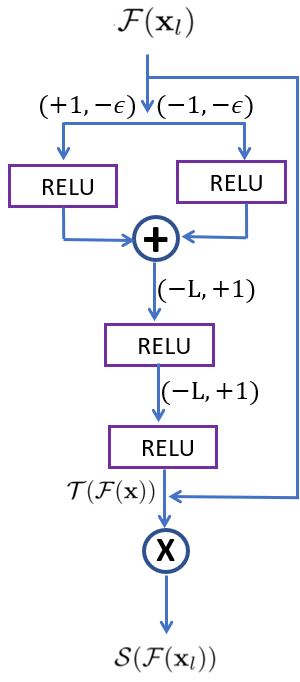,width=0.60\columnwidth}\label{fg.epsilon_resnet_design.c}}
}
\end{center}
\caption{(a) We show the network structure of one of the residual blocks in the standard pre-activation network~\cite{he2016identity}. (b) We show the network structure of one of the residual blocks in $\epsilon$-ResNet. We have added a sparsity-promoting function ${\cal S}()$ to discard the residual if all the individual responses are less than a threshold $\epsilon$. (c) We show the network structure for the sparsity-promoting function ${\cal S}()$ using 4 RELU's and one multiplicative gate. The pair $(i,j)$ shown in brackets denote the weights $i$ and bias term $j$ for the associated RELU function. $L$ refers to a very large positive constant and $\epsilon$ denotes the constant that we want to use for discarding layers.}
\label{fg.epsilon_resnet_design}
\end{figure*}

\section{$\epsilon$-ResNet}
\noindent
{\bf Standard ResNet:} Residual network consists of a large stack of residual blocks. Each residual block has the architecture shown in Fig.~\ref{fg.epsilon_resnet_design.a}. Each residual block can be seen as the following mapping:
\begin{equation}
{\cal H}({\bf x}) = {\bf x} + {\cal F}({\bf x})
\end{equation}

The residual block consists of pre-activations of weight layers as proposed in the improved version of residual networks~\cite{he2016identity}. Note that this is in contrast to the earlier version that used post-activations (RELU and Batch normalization BN) of weight layers~\cite{he2016deep}. In all our experiments, we use the pre-activation residual network as the baseline. Our proposed $\epsilon$-ResNet will also be built on top of this pre-activation residual networks. The entire network is built by stacking multiple residual blocks. It has three groups of residual blocks, and each group has equal numbers of blocks. For example, in the case of a 110-layer network, we have 3 groups each having 18 blocks. Each block has 2 layers, and thus the three groups will have a total of 108 ($3 \times 2 \times 18$) layers. In addition, we have one convolution layer before and one fully connected layer after the three groups of residual blocks, and thereby leading to 110 layers. The dimension of the first group is 16 and is multiplied by two in the later two groups, and the feature-map sizes in the three groups are $32\times32, 16\times16, 8\times8$.

\noindent
{\bf $\epsilon$-ResNet:}
Fig.~\ref{fg.epsilon_resnet_design.b} shows the mapping function in every block in the $\epsilon$-ResNet. The basic idea in $\epsilon$-ResNet is to use a {\bf sparsity-promoting function} ${\cal S}({\cal F}({\bf x}))$ that automatically discards the residual ${\cal F}({\bf x})$ if all the individual responses are less than a threshold $\epsilon$. Instead using a function mapping ${\cal H}({\bf x}) = {\bf x} + {\cal F}({\bf x})$, we use a function mapping as shown below:

\begin{equation}
{\cal H}({\bf x}) = {\bf x} + {\cal S}({\cal F}({\bf x}))
\end{equation}

Let us assume that ${\cal F}({\bf x})$ is a vector of length $n$. Let each element of the vector is denoted by ${\cal F}({\bf x})_i$ where $i \in \{1,\dots,n\}$. The sparsity-promoting function is defined below:
\begin{equation}
{\cal S}({\cal F}({\bf x})) =
\begin{cases}
    0 & \mbox{if $|{\cal F}({\bf x})_i| < \epsilon,\forall i \in \{1,\dots,n\}$}\\
    {\cal F}({\bf x}) & \mbox{otherwise.}
  \end{cases}
\end{equation}

\noindent
{\bf Proposed Structure:} Fig.~\ref{fg.epsilon_resnet_design.c} shows our proposed network structure to implement the sparsity promoting function ${\cal S}({\cal F}({\bf x}))$. The network consists of 4 RELU layers and one multiplicative gating function $(\times)$, as utilized in ~\cite{srivastava2015highway}.

Let us study the behaviour of this network. First, let us consider the case when all the response elements satisfy the condition $|{\cal F}({\bf x})_i| < \epsilon$. In this case, the output from the summation ($+$) will be zero. A zero input to the third RELU will lead to an output of $1$. An output of $1$ from the third RELU will lead to an output of $0$ from the fourth RELU. Finally, we will have ${\cal T}({\cal F}({\bf x})) = 0$, and thus we have the following:
\begin{equation}
{\cal S}({\cal F}({\bf x})) = {\cal T}({\cal F}({\bf x})) \times {\cal F}({\bf x})
= 0
\end{equation}

Now let us consider the second scenario where at least one of the response elements satisfies either ${\cal F}({\bf x}) > \epsilon$ or ${\cal F}({\bf x}) < -\epsilon$. In this case we will have a non-zero positive output from the first summation $(+)$. Non-zero positive output from the summation would lead to $0$ output after the third RELU, and eventually output ${\cal T}({\cal F}({\bf x})) = 1$ from the final RELU. Thus we have:
\begin{equation}
{\cal S}({\cal F}({\bf x})) = {\cal T}({\cal F}({\bf x})) \times {\cal F}({\bf x})
= {\cal F}({\bf x})
\end{equation}\newline

\noindent
{\bf Loss function:}
We use the same loss function that was used in ResNet~\cite{he2016identity} involving the cross-entropy term and L2 regularization on the weights. Note that we also use a side-supervision in addition to the loss function at the last output layer. Let us denote the additional function used for side-supervision as the side loss. If there are N residual blocks, the output of [N/2]-th block is used in the computation of the side loss. To get side loss, we follow the same architecture as the standard loss function. We apply a fully-connected layer before the soft-max layer and then forward their output to the cross entropy function. The output dimension of the fully-connected layer is the number of classes in the dataset. Finally, side loss is used in the overall cost function with a coefficient of 0.1. Side supervision can be seen as a strategy to shorten the path of back propagation for the first half of layers during training. Note that side supervision is not involved in prediction.
\newline

\begin{figure}[!htbp]
\begin{center}
\psfig{figure=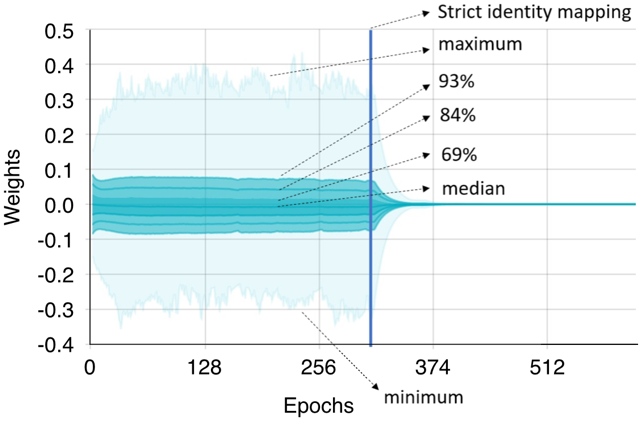,width=1.00\columnwidth}
\end{center}
\caption{We show the histogram of all weights in one of the layers of a residual block that achieves strict identity. The $x$-axis shows the index of the epochs. The $y$-axis shows the value of the weights. The different curves show the maximum, minimum, 93rd percentile, etc. As we observe, the weights collapse to zeros once a residual block is identified as unnecessary by our$\epsilon$-ResNet algorithm.}
\label{fg.weight_collapse}
\end{figure}

\noindent
{\bf Weight collapse:}
When a residual block produces negligible responses, the sparsity promoting function will start producing $0$ outputs. As a result, the weights in this block will stop contributing to the cross-entropy term. Consequently the gradients will be only based on the regularization term, and thus the weights in the associated residual block will move to $0$'s. Note that the weights don't go to zeros instantly, and the number of iterations necessary for reaching zeros depends on the learning rate and momentum parameters. In Fig.~\ref{fg.weight_collapse}, we show the weight collapses for one of the layers in a residual block that starts to model strict identity mapping.

\section{Experiments}
We evaluate $\epsilon$-ResNet on four standard datasets: CIFAR-10 and CIFAR-100~\cite{krizhevsky2009learning}, SVHN~\cite{SVHN}, and ImageNet 2012 dataset~\cite{imagenet}. We used standard ResNet as the baseline for comparison. \newline

\begin{figure*}[!htbp]
\begin{center}
\mbox{
\subfigure[]{\psfig{figure=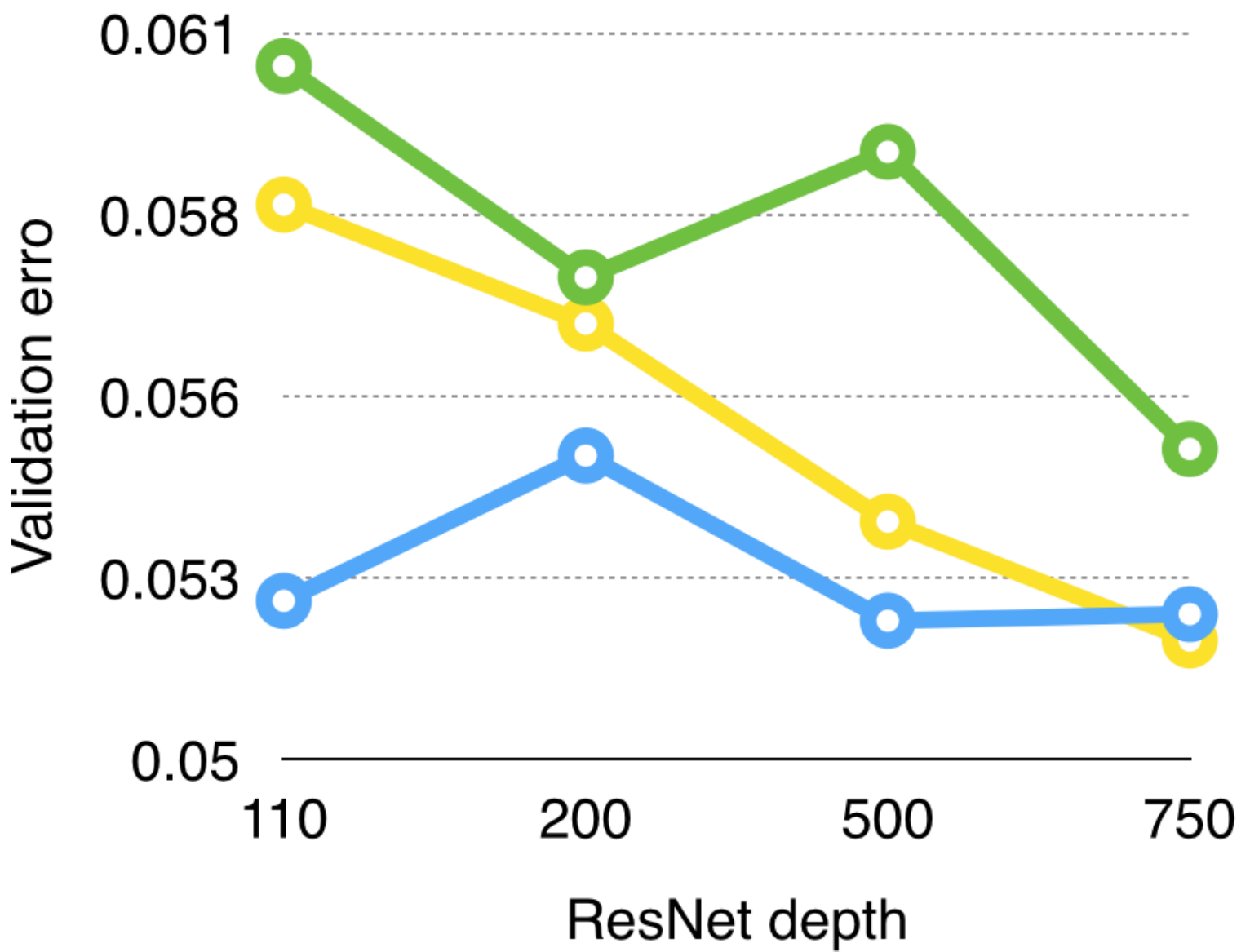,height=3.5cm}}
\subfigure[]{\psfig{figure=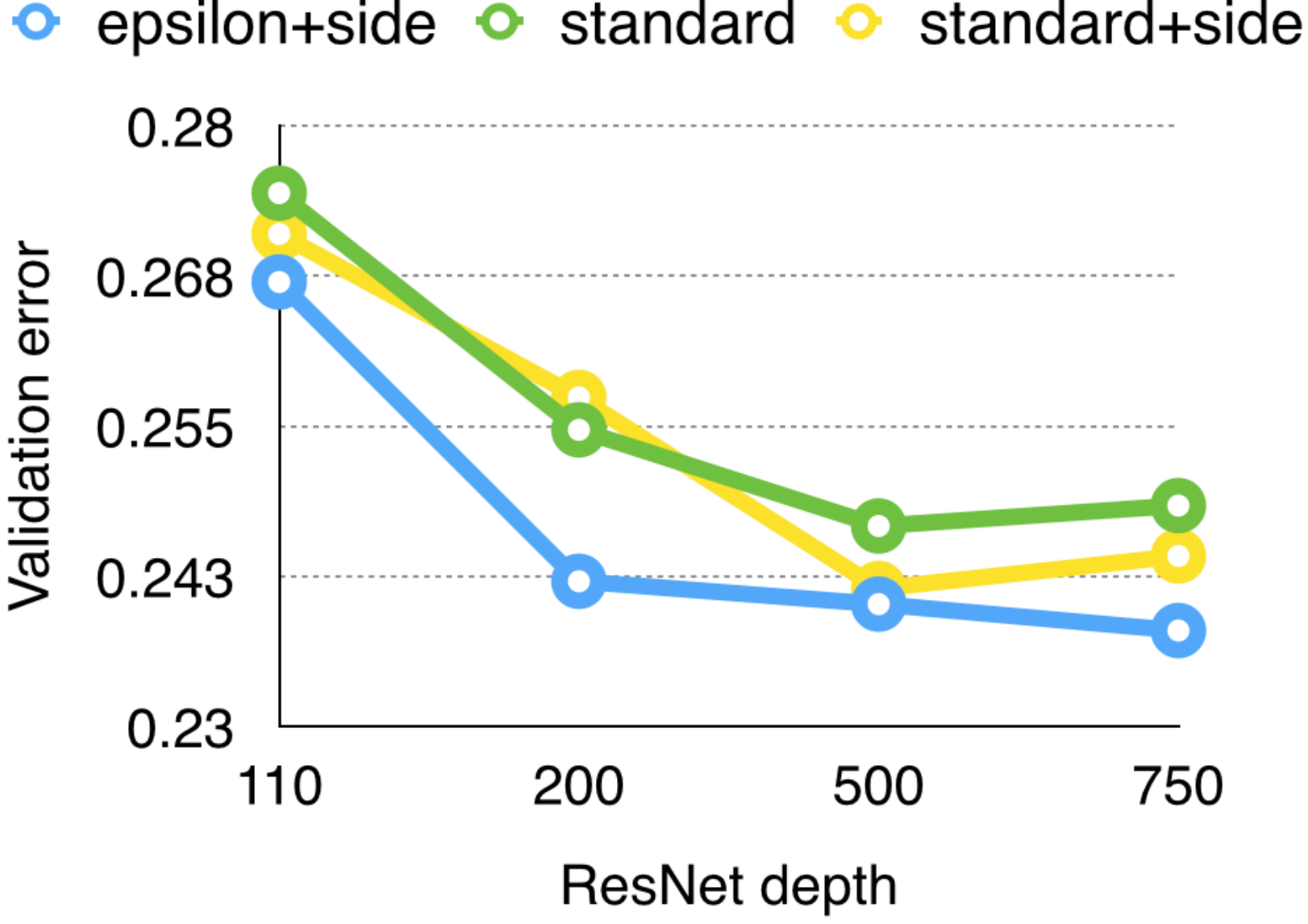,height=3.5cm}}
\subfigure[]{\psfig{figure=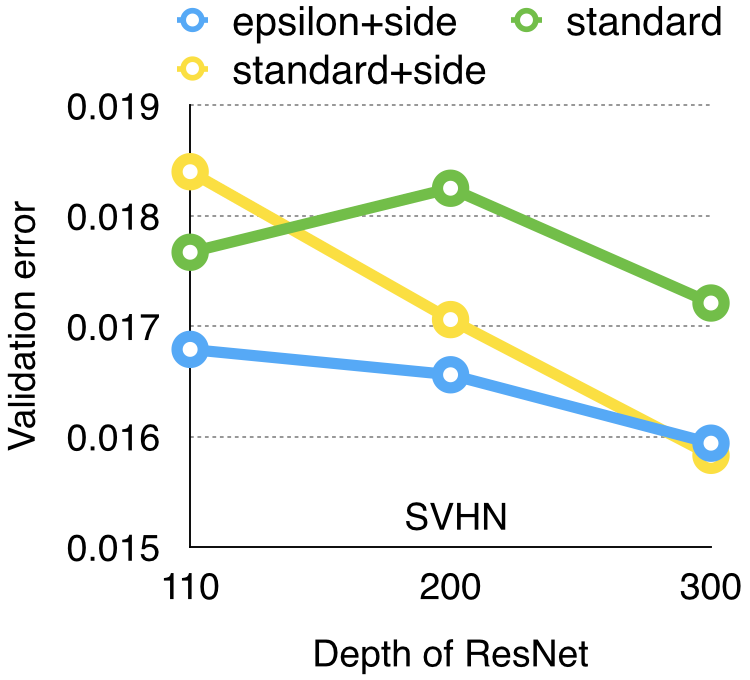,height=3.5cm}}
\subfigure[]{\psfig{figure=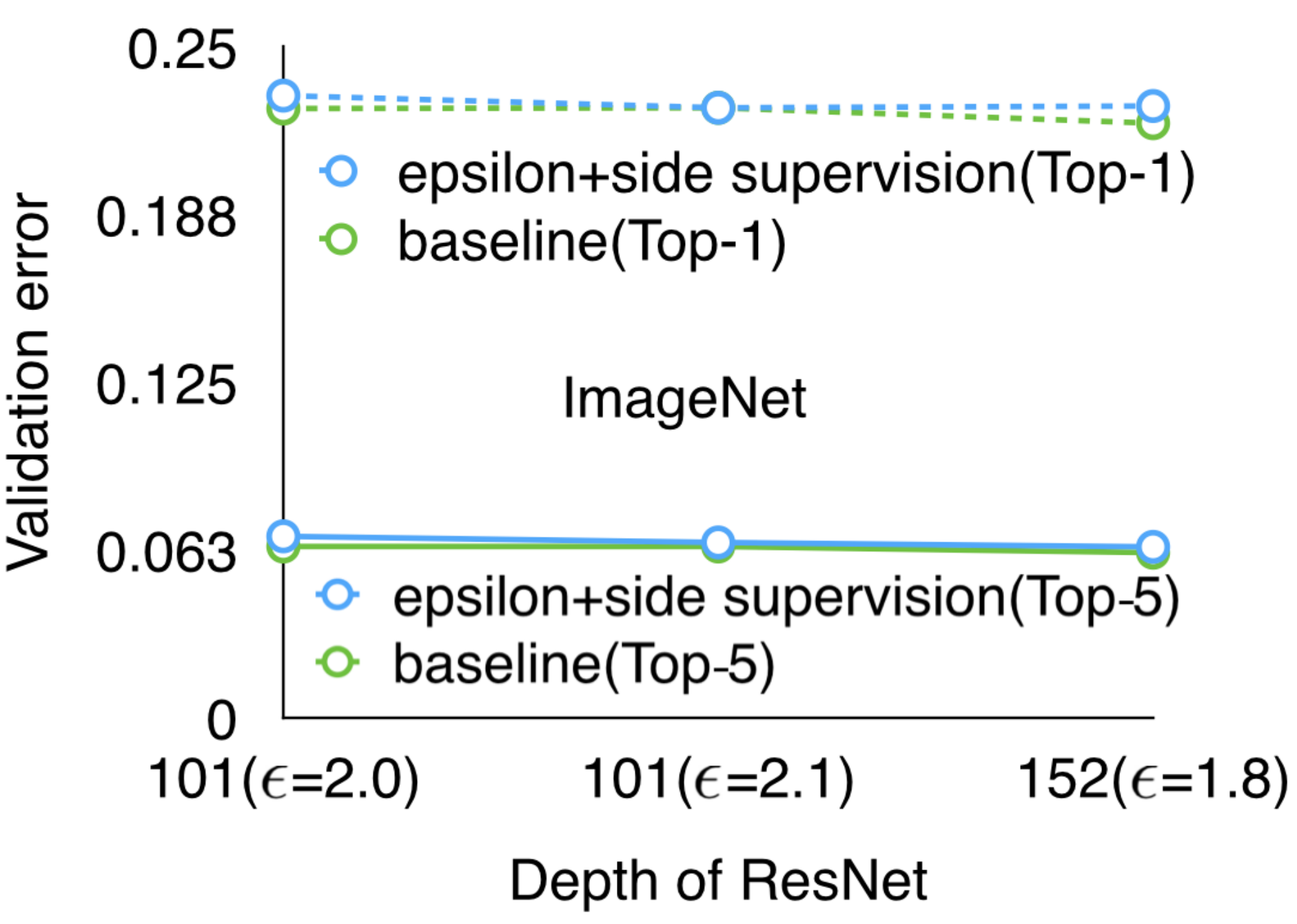,height=3.5cm}}
}
\mbox{
\subfigure[]{\psfig{figure=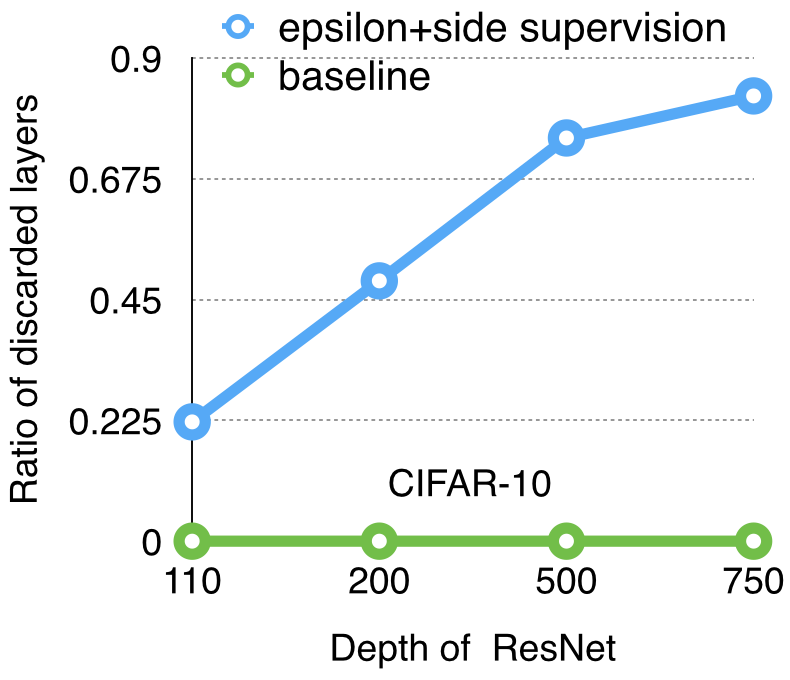,height=3.3cm}}
\subfigure[]{\psfig{figure=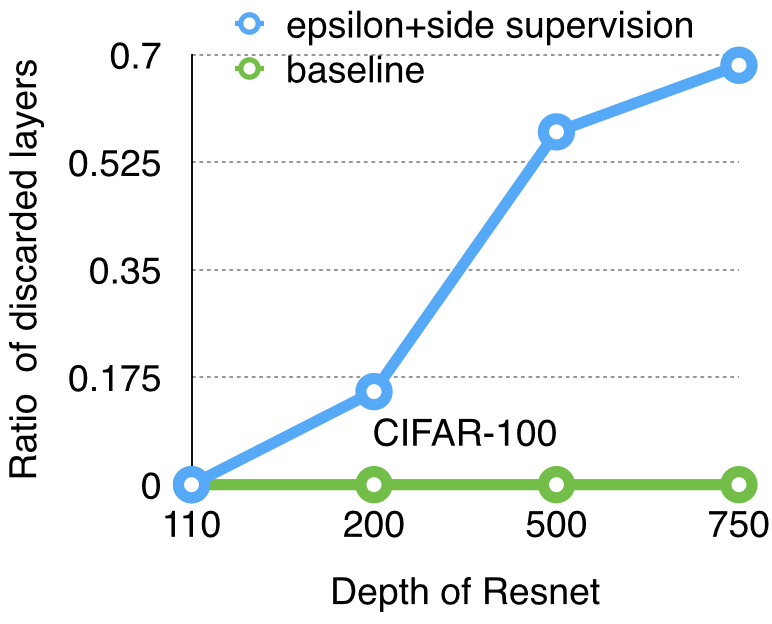,height=3.3cm}}
\subfigure[]{\psfig{figure=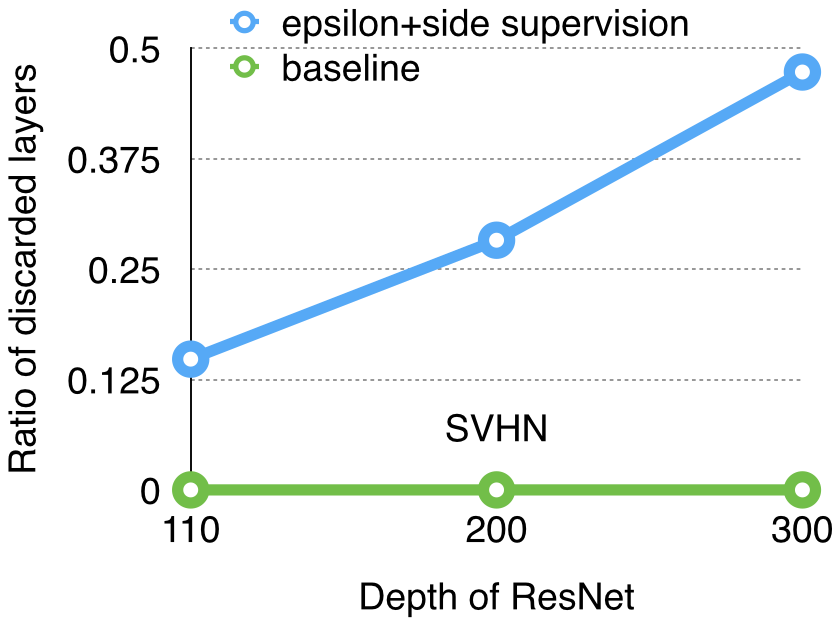,height=3.3cm}}
\subfigure[]{\psfig{figure=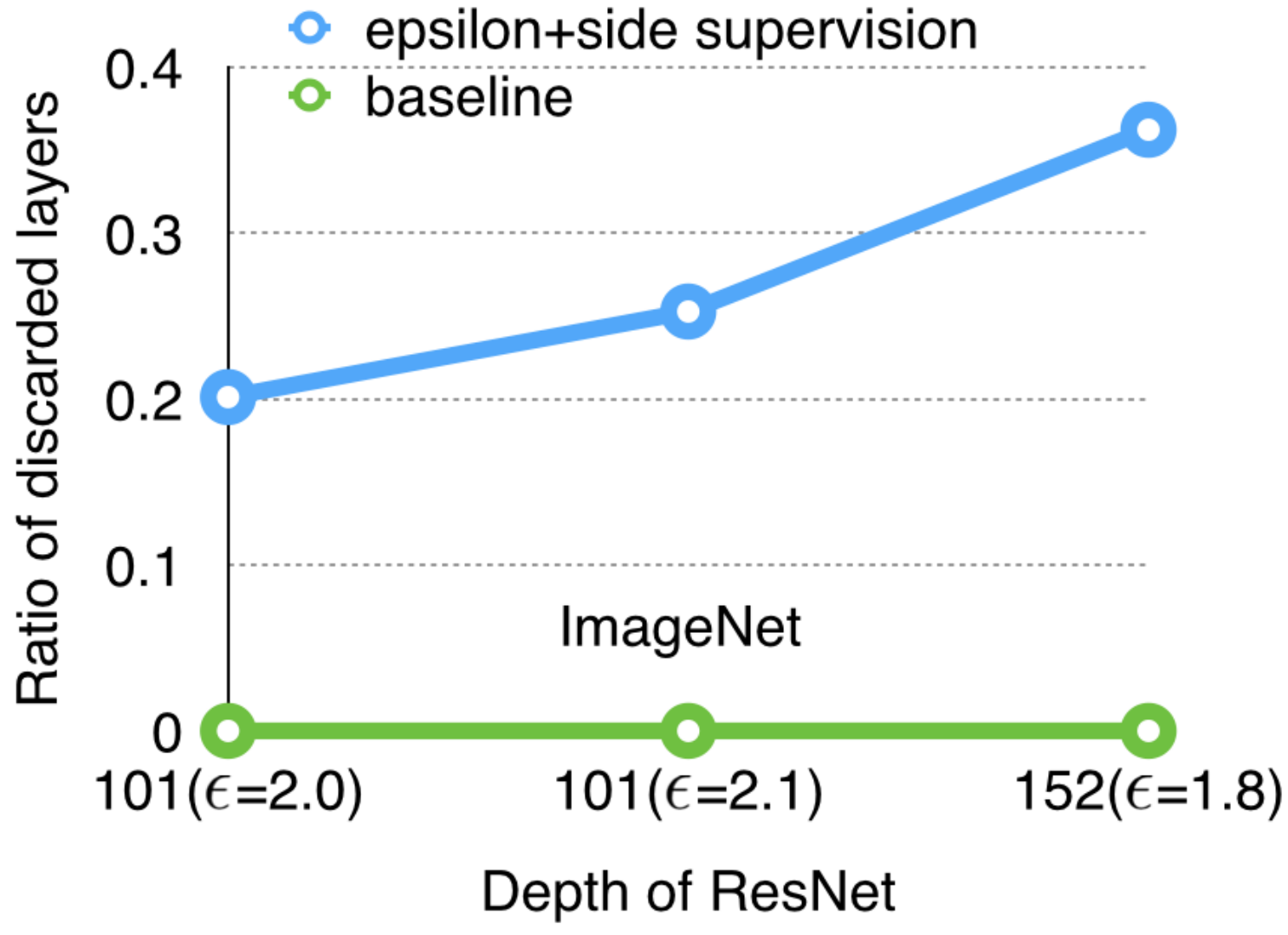,height=3.3cm}}
}
\end{center}
\caption{(a), (b), (c), and (d) show the validation error of ResNet and $\epsilon$-ResNet with different number of layers on CIFAR-10, CIFAR-100, SVHN, and ImageNet, respectively. (e), (f), (g), and (h) show the ratio of discarded layers of $\epsilon$-ResNet with a different number of layers on CIFAR-10, CIFAR-100, SVNH, and ImageNet, respectively. The validation error of ResNet-152 baseline is borrowed from https://github.com/facebook/fb.resnet.torch.}
\label{fg.errors_and_ratios}
\end{figure*}

\begin{figure*}[!htbp]
\begin{center}
\mbox{
\subfigure[]{\psfig{figure=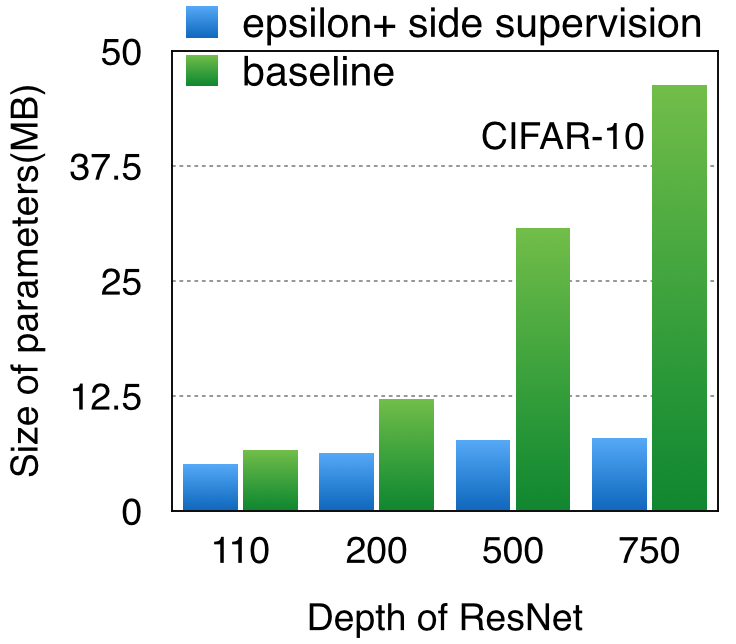,height=3.7cm}}
\subfigure[]{\psfig{figure=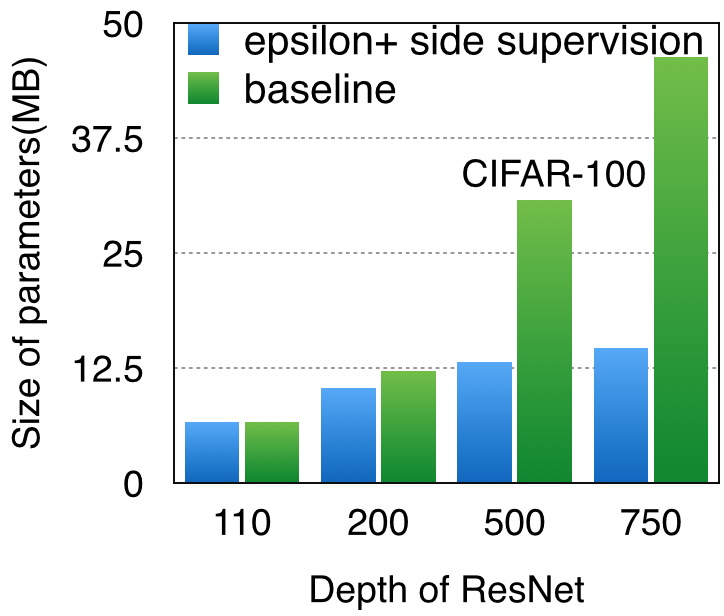,height=3.7cm}}
\subfigure[]{\psfig{figure=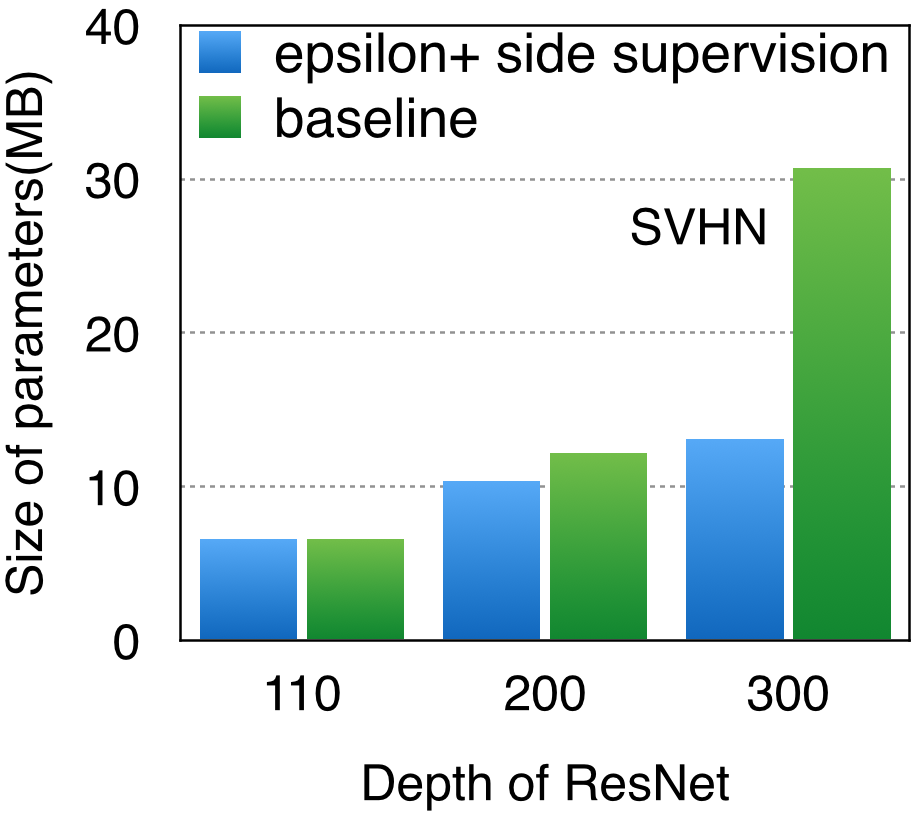,height=3.7cm}}
\subfigure[]{\psfig{figure=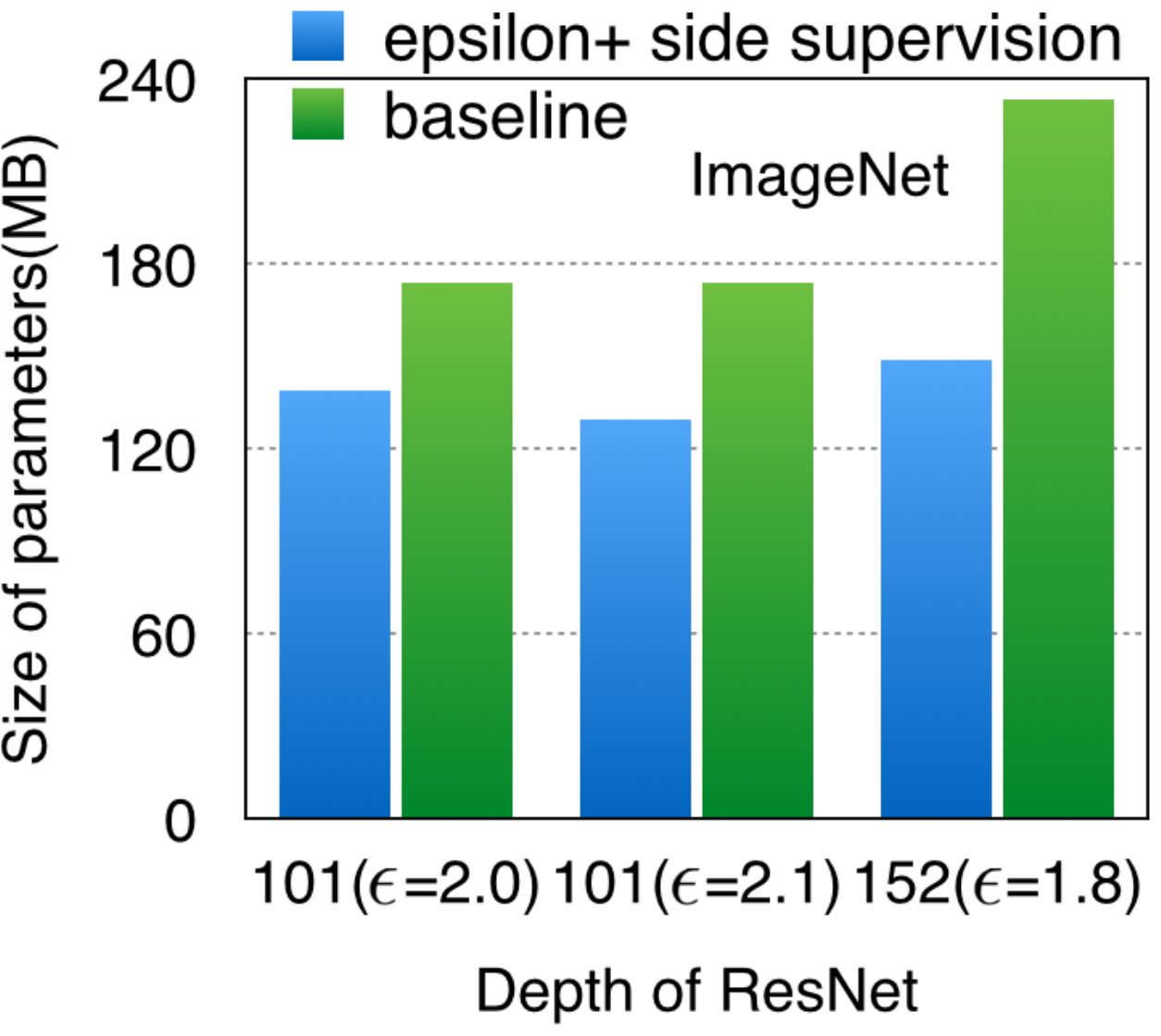,height=3.7cm}}
}
\end{center}
\caption{(a), (b), (c), and (d) show the memory consumption for parameters used in the standard ResNet and the reduced one for CIFAR-10, CIFAR-100, SVHN, and ImageNet experiments, respectively. As we can see, we achieve a significant compression in the case of networks with a large number of layers. 
}
\label{fg.compression_and_speedup}
\end{figure*}

\begin{figure}[!htbp]
\begin{center}
\psfig{figure=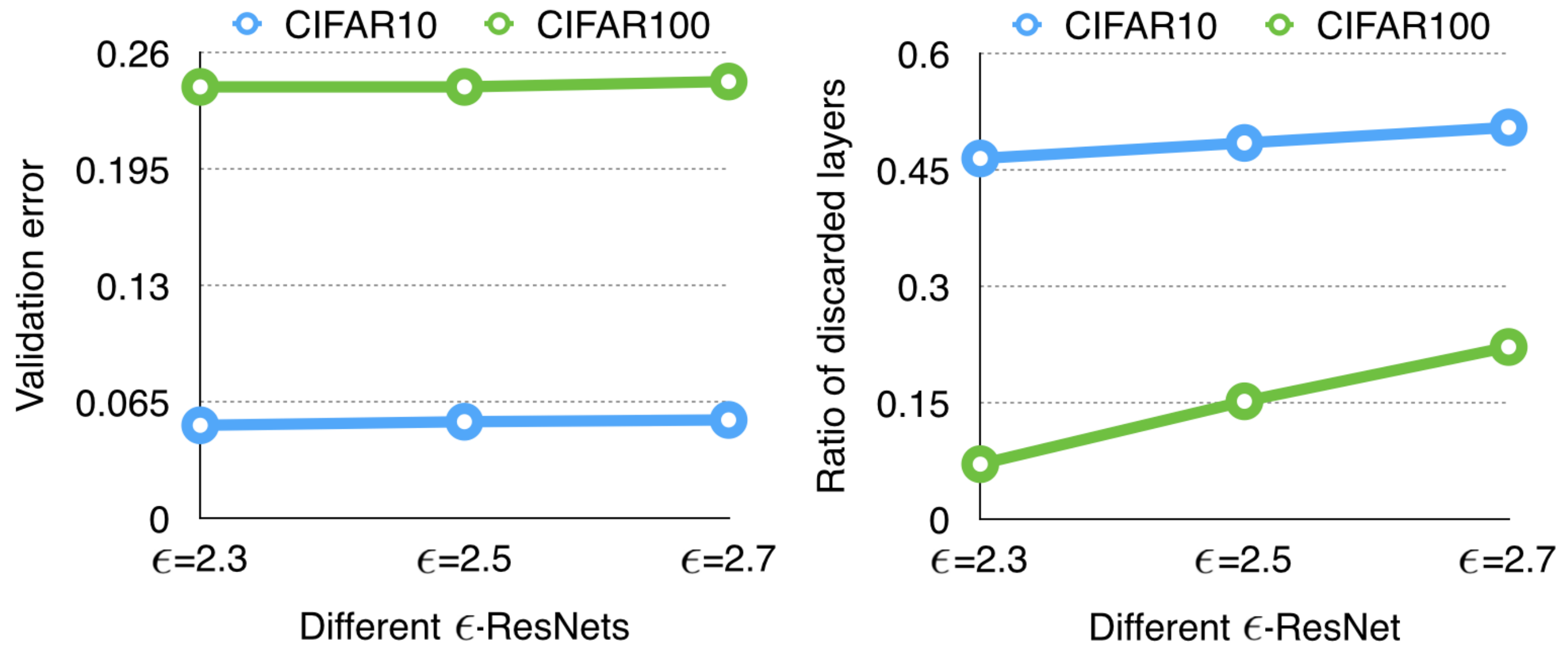
,width=0.95\columnwidth}
\end{center}
\caption{Sensitivity to the choice of $\epsilon$ parameter values: We studied the ratio of discarded layers and validation errors for different $\epsilon$ parameter values in the case of CIFAR-10 and CIFAR-100. We observed that $\epsilon$ parameter can be seen as a tuning parameter to achieve the tradeoff between compression and accuracy.}
\label{fg.epsilon_sensitivity}
\end{figure}

\begin{figure}[!htbp]
\begin{center}
\psfig{figure=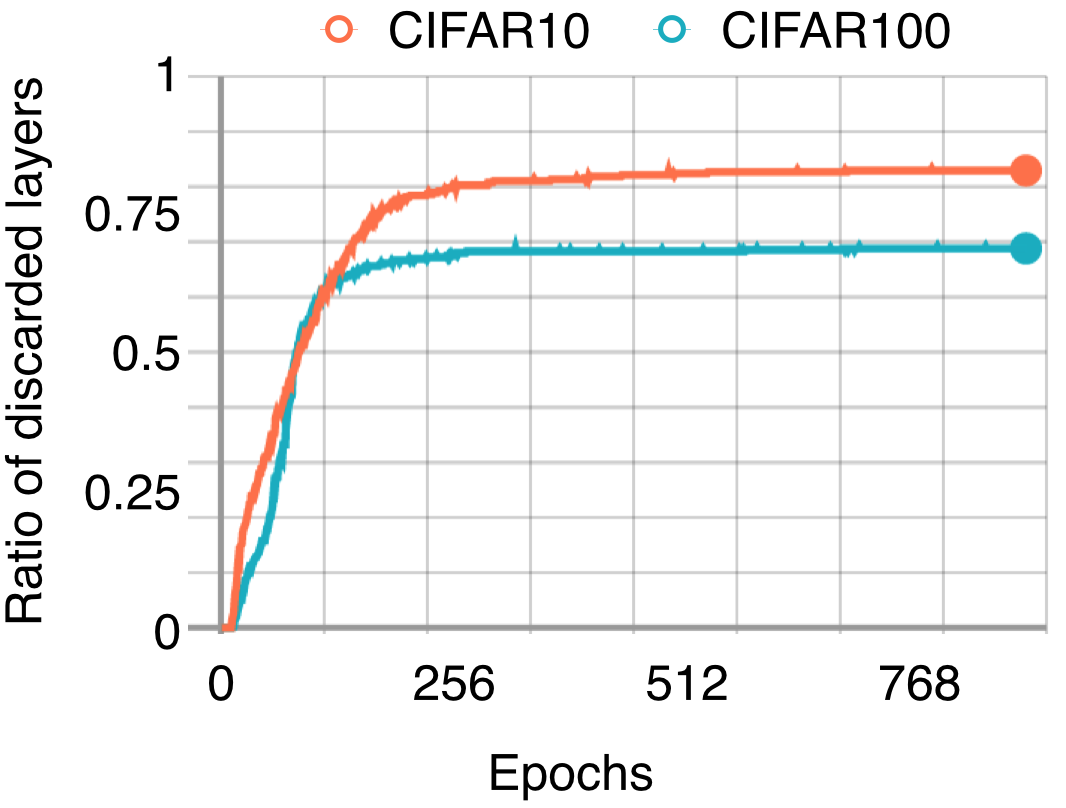,width=0.90\columnwidth}
\end{center}
\caption{We study the ratio of discarded layers with respect to the number of epochs. Results are shown for a 752-layer $\epsilon$-ResNet on CIFAR-10 and CIFAR-100.}
\label{fg.str_ratio_with_iterations}
\end{figure}

\begin{figure}[!htbp]
\begin{center}
\mbox{
\subfigure[]{\psfig{figure=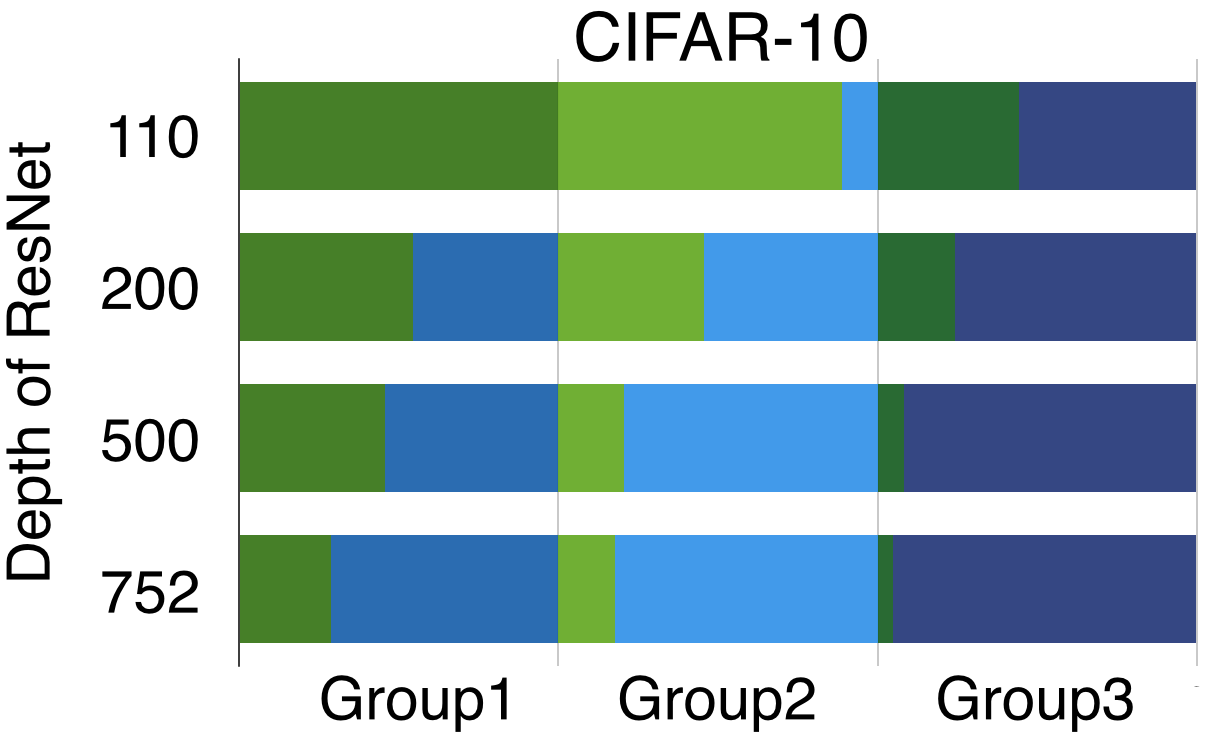,width=0.50\columnwidth}}
\subfigure[]{\psfig{figure=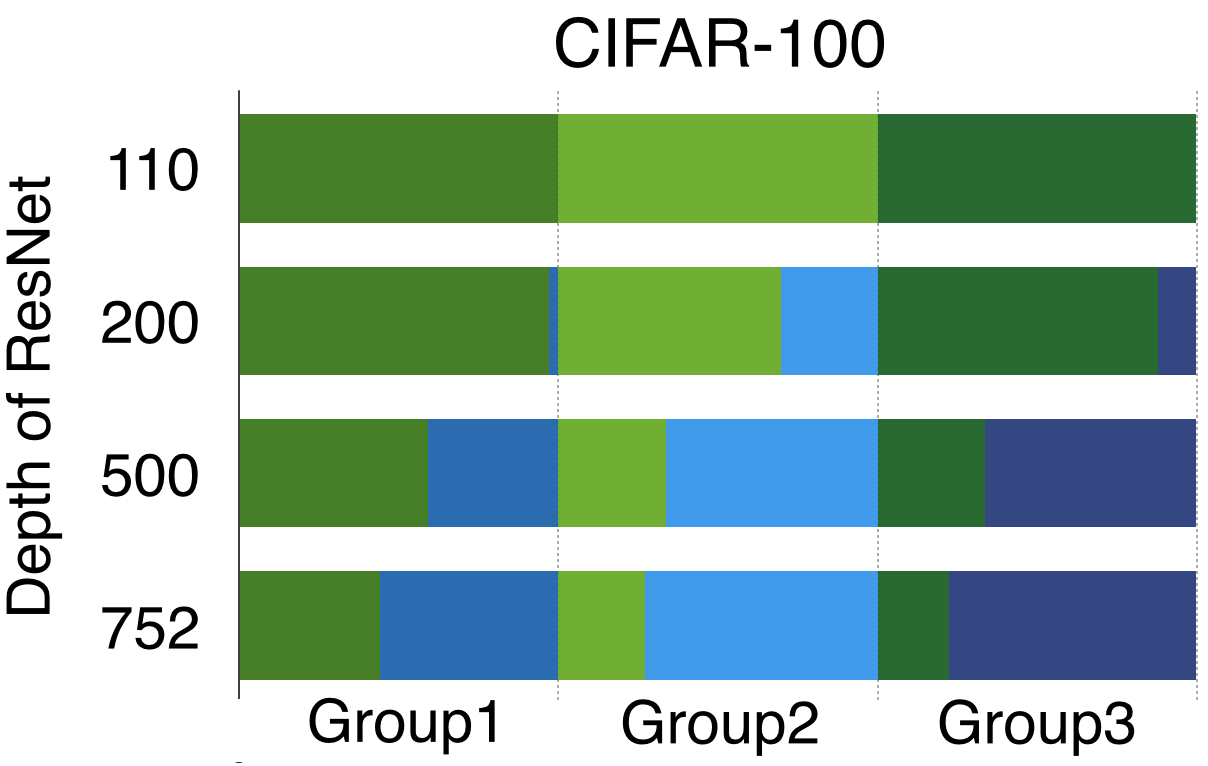,width=0.50\columnwidth}}
}
\end{center}
\caption{Group-wise proportions of layers that are not discarded versus the discarded ones on ResNets with different depths. All network structures in the experiment contain three groups, where each group contains layers at the same resolution. For each group, green bar represents the proportion of layers not discarded, while blue bar represents the proportion of discarded layers. (a) and (b) show the results on CIFAR-10 and CIFAR-100, respectively.}
\label{fg.groupwise_compression}
\end{figure}

\noindent
{\bf Datasets:} The CIFAR-10 dataset~\cite{krizhevsky2009learning} consists of $60,000$ $32 \times 32$ RGB images with 10 classes, where each class has $6,000$ images. The CIFAR-100 dataset~\cite{krizhevsky2009learning} also consists of $60,000$ $32 \times 32$ RGB images with 100 classes, where each class has 600 images. We use $50,000$ training images and $10,000$ test images in both the datasets. We follow the standard data augmentation that allows for small translations and horizontal flips. We first pad 4 pixels on all sides, then we randomly crop $32 \times 32$ images from the padded image, and finally we do a horizontal flip. \newline

In the Street View House Numbers (SVHN)~\cite{SVHN} dataset, the models are trained on $604,388$ RGB images and evaluated on $26,032$ RGB images of size $32\times32$. For data augmentation, we apply the same augmentation method as on the CIFAR datasets. Before random cropping, we apply the following additional transformations to augment the data: For every training image, we perform scaling and transformation with one random factor on all its pixels. In particular, we add a random number in $[-10, 10]$ for all pixels on one image, then for each pixel randomly multiply the residual between RGB channel values and their channel mean with a scale between $[0.8, 1.2]$, add the scaled residuals to the original mean. The randomly scaled and transformed values are then truncated to the range of $[0, 255]$ before output.\newline

The ImageNet 2012 classification dataset~\cite{imagenet} has $1,000$ classes and contains $1,281,167$ training images and $50,000$ validation images. We report both top-1 and top-5 validation error rates on it. We follow the practice in ~\cite{he2016deep, DBLP:journals/corr/Howard13, szegedy2015going,NIPS2012_4824} to conduct data augmentation on ImageNet. A $224\times 224$ crop is randomly sampled from an image by following the scale and aspect ratio augmentation from ~\cite{szegedy2015going}, instead of scale augmentation used in the ResNet~\cite{he2016deep} paper because the former gives a better validation error. We then apply the following transformations in a random order: horizontal flip, the standard AlexNet-style color augmentation~\cite{NIPS2012_4824} and photometric distortions~\cite{DBLP:journals/corr/Howard13}.
\newline

\noindent
{\bf $\epsilon$ Parameter:}
We can think of $\epsilon$ as a hyperparameter that allows us to find a tradeoff between accuracy and network size. In practice, identifying a good $\epsilon$ parameter is not difficult due to the following reasons: (1) The $\epsilon$ values are generally concentrated within a small range to produce good results and compression, as shown in the experiments. (2) $\epsilon$ satisfies nice monotonicity property (larger value leads to higher compression). (3) In all the experiments (CIFAR-10, CIFAR-100, SVHN, and Imagenet), we were able to quickly find $\epsilon$ parameters (generally in the range 1.5-3 with just 1 or 2 attempts). For example, the experiments for CIFAR-10 and CIFAR-100 used an $\epsilon$ of $2.5$. For all experiments on SVHN, we use an $\epsilon$ of 1.5. We use $\epsilon$ in the range (1.8, 2.1) on ImageNet. Fig.~\ref{fg.epsilon_sensitivity} shows that in a reasonable range, greater $\epsilon$ brings larger compression ratio.
\newline

\noindent
{\bf Number of layers:} We experimented with networks of depths 110, 200, 500, and 752-layers on CIFAR-10 and CIFAR-100. These four networks have 54, 99, 249, 375 residual blocks (consisting of two $3\times3$ convolutional layers each) respectively. Considering the size of SVHN is 10 times larger than CIFAR-10, we test SVHN with networks of depths 100, 200, and 302 layers. Finally, we evaluated the models of depths 101 and 152 on ImageNet. \newline

\noindent
{\bf Adaptive learning rate:}
Our learning rate scheduling closely follows the standard ResNet implementation~\cite{he2016deep}. On CIFAR dataset, we start with a learning rate of 0.1 and decrease it by a factor of 10 at epochs 82 and 123. On SVHN dataset, we begin with 0.1 and decrease it by a factor of 10 at epochs 20, 28 and 50. When the network starts to lose layers, we will stop using the standard learning rate policy and start using adaptive learning rate policy. According to our adaptive learning rate policy, every time we lose a residual block, we reset the learning rate to the initial value of the standard setting and decrease it at a rate that is twice as fast as the standard policy. In other words, we decrease it by a factor of 10 after 41 and 61 epochs for CIFAR datasets. For SVHN, we will start decreasing by a factor 10 after 10, 14, and 25 epochs. Such adaptive learning rate policies have been used before~\cite{Smith2017CyclicalLR,Schaul2013NoMP}. 

Adaptive learning rate policy was not necessary for ImageNet, where we just start with a learning rate of 0.1 and decrease it by a factor of 10 at epochs 30, 60, 85, and 95.\newline

\noindent
{\bf Training:}
We implement $\epsilon$-ResNet using Tensorflow on TITAN XP and GeForce GTX 970 graphic cards. Our code is built upon the tensorpack~\cite{wu2016tensorpack}, an implementation of ResNet can be found \href{https://github.com/ppwwyyxx/tensorpack/tree/master/examples/ResNet}{here}. We followed the standard Gaussian initialization of 0 mean and 0.01 std for weights, and constant initialization of 0s for biases. 

Following the practice in~\cite{he2016deep}, we use standard data augmentation methods and train the network using SGD with a mini-batch size of 128 on CIFAR and SVHN datasets and a mini-batch size 256 on ImageNet for each GPU. On CIFAR datasets the baseline models are trained for up to 200 epochs ($78,000$ iterations), while $\epsilon$-ResNets requires $1,000$ epochs to train since it uses adaptive learning rate. For SVHN, the baseline and $\epsilon$-Resnet used 80 and 150 epochs, respectively. All the models on ImageNet are trained for 110 epochs. We use a weight decay of 0.0002 and a momentum of 0.9 for all the experiments.\newline

\noindent
{\bf Side-supervision:}
Side-supervision is a strategy to impose additional losses at intermediate hidden layers in addition to the loss on the top~\cite{Xie2015HolisticallyNestedED}. In $\epsilon$-ResNet, we apply one additional loss at the middle of the network with a coefficient of 0.1. We observed that side supervision allows the weights to generally decrease as we move from the input layer to the output layer. As shown in Fig.~\ref{intro_figure}, we reject more layers near the output layer as shown by the red lines.

In Fig.~\ref{fg.errors_and_ratios}, we also show the results for the standard Resnet improves with side-supervision. However, $\epsilon$-ResNet still achieves similar performance, along with providing the additional benefit of significant compression.\newline

\noindent
{\bf Validation errors:} Fig.~\ref{fg.errors_and_ratios} report the comparison in validation errors between standard ResNet and $\epsilon$-ResNet. $\epsilon$-ResNet again achieves a significant reduction in model size while maintaining good prediction performance on CIFAR-10, CIFAR-100, and SVHN. On these three datasets, we did not find any degradation in the performance even after discarding a significant number of layers. We also evaluated the performance of $\epsilon$-ResNet with 101 and 152 layers on ImageNet. With a marginal loss in performance, $\epsilon$-ResNet discarded 20.12\%,  25.60\%, and 36.23\% of layers for different $\epsilon$ parameters and depths.
\newline

\noindent
{\bf Memory consumption:}
The memory footprint reduction for CIFAR-10, CIFAR-100, SVHN, and ImageNet are shown in Fig.~\ref{fg.compression_and_speedup}, respectively.  \newline

\noindent
{\bf Layer selection}
Fig.~\ref{fg.str_ratio_with_iterations} shows that the proportion of strict identity mapping increases with the training iterations and gradually saturates, finally stabilizing at a ratio where the performance is similar to the original. When looking into the final learned structure, Fig.~\ref{fg.groupwise_compression} shows that $\epsilon$-ResNet is prone to discarding more layers closer to the output. \newline

\section{Discussion}
We propose $\epsilon$-ResNet, a variant of standard residual networks~\cite{he2016identity}, that automatically identifies and discards redundant layers with marginal or no loss in performance. We achieve this using a novel architecture that enables strict identity mappings when all the individual responses from a layer are smaller than a threshold $\epsilon$. We tested the proposed architecture on four datasets: CIFAR-10, CIFAR-100, SVHN, and ImageNet. We plan to explore two avenues in future. First, we will focus on developing an algorithm that can automatically identify a good $\epsilon$ value that produces maximum compression with marginal or no loss in performance. Second, we will extend the proposed method to architectures beyond ResNet. 

\section*{Acknowledgments}
\noindent
We thank the reviewers and area chairs for valuable feedback. S. Ramalingam would like to thank Mitsubishi Electric Research Laboratories (MERL) for partial support. We thank Tolga Tazdizen and Vivek Srikumar for GPU resources.
{\small
\bibliographystyle{ieee}
\bibliography{ref}
}

\end{document}